# PEAK DETECTION ON DATA INDEPENDENT ACQUISITION MASS SPECTROMETRY DATA WITH SEMISUPERVISED CONVOLUTIONAL TRANSFORMERS

*Leon L. Xu, Hannes L. Röst*

University of Toronto

**ABSTRACT**

Liquid Chromatography coupled to Mass Spectrometry (LC-MS) based methods are commonly used for high-throughput, quantitative measurements of the proteome (i.e. all proteins in a sample). Targeted LC-MS produces data in the form of a two-dimensional time series spectrum, with the mass to charge ratio of analytes (m/z) on one axis, and the retention time from the chromatography on the other. The elution of a peptide of interest produces highly specific patterns, called peaks, across multiple fragment ion traces (extracted ion chromatograms, or XICs). In this paper, we formulate this peak detection problem as a multi-channel time series segmentation problem, and propose a novel approach based on the Transformer architecture. Here we augment Transformers, which can capture global interactions, with Convolutional Neural Networks (CNNs), which can capture local context, in the form of Transformers with Convolutional Self-Attention. We explore how to train this model in a semisupervised manner by adapting a state-of-the-art semisupervised image classification technique to multi-channel time series data. Experiments on a representative LC-MS dataset are benchmarked using manual annotations to showcase the encouraging performance of our method; it outperforms baseline neural network architectures and is competitive against the current state-of-the-art in automated peak detection.

## 1. INTRODUCTION

Mass spectrometry (MS) is an analytical technique which measures the abundance of analytes of interest (e.g. short peptides in the context of proteomics) sorted by their mass-to-charge ratios (m/z). Chromatography is a means of analytical separation commonly used as a sample preparation step before performing mass spectrometry that feeds separate parts of an input mixture based on physical properties of the contained molecules into the mass spectrometer in sequence. Coupling these two methods together "on-line" produces two-dimensional time series spectra, with the m/z of analytes on one axis and the time the analyte elutes from the chromatography to the mass spectrometer (retention time, or RT) on the other (Fig. 1 (a)).

In a standard proteomics LC-MS experiment, a complex mixture of proteins is enzymatically or chemically digested into short peptides of approximately 7 to 30 amino acids in length, which are then separated by chromatography and injected into a mass spectrometer. The MS instrument acquires spectra of the whole peptides (MS1 spectra) and then transfers energy to the peptides to produce fragment ions which are recorded in so-called MS2 spectra. During data analysis, these signals need to be associated with the peptide that produced them for accurate identification and quantification. In Data Independent Acquisition (DIA) data analysis, a specific paradigm for the analysis of mass spectrometric data, prior information about a peptide precursor is used to extract a number of time series signals from the m/z axis to form an n-channel time series corresponding to the precursor and its corresponding fragments (an extracted ion chromatogram, or XIC) [12]. However, a sample may or may not actually contain a peptide and due to experimental variation in chromatographic elution, the true location of the peak signal along the RT axis is also unknown. The primary challenge in peak detection is being able to differentiate between true peptide signals and false noise signals, which can be due to chemical noise, instrument noise, or partial true signals from similarly fragmented peptides, among other sources. Peak signals can also occur across an extremely large dynamic range of intensity values, making it very difficult to detect low abundance signals among relatively high noise using standard heuristics-based thresholding or filtering techniques. Finally, as mass spectrometry technology improves, more information that can potentially be used to aid in the peak detection task is available, but it is extremely difficult for existing knowledge-based software to incorporate the increasingly complex resulting data.

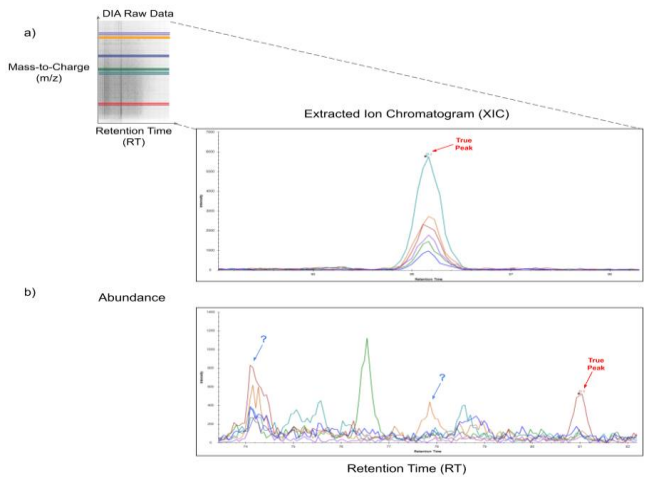

**Figure 1:** a) Individual traces are extracted from the raw data to generate individual XICs. b) An example of a more difficult to annotate case. The x-axis is the retention time of analytes in the chromatography, and the y-axis is the abundance of analytes.

Current go-to approaches generally feature a three step process, after generating an n-channel XIC - one channel for each of a precursor's fragments: first, peaks are selected from individual channels of the XIC by simply selecting local maxima with decreasing intensity on both sides, then peak region candidates are created by finding overlapping peaks across other channels, and finally manually engineered features based on characteristics of the peak region candidates (such as shape correlation of peaks across channels) are computed and fed as input into a machine learning model, such as LDA or XGBoost [10]. Deep learning has already been applied towards the peak detection problem in the form of standard Feed Forward Networks in [8], using them as a drop-in replacement for the machine learning methods in the standard peak detection pipeline. There has also been work using deep learning to refine the boundaries of confirmed peak signals as presented in [19], but the method cannot distinguish true signal from noise and is thus only applicable to short stretches of the overall time series once detection and localization have already been performed, and the method currently only operates on a single channel and cannot exploit the multi-channel information of real-world datasets.

Here we propose an end-to-end peptide peak detection system which uses Transformers with Convolutional Self-Attention that can operate directly on more complex XICs. We show that our model can be trained successfully in a semisupervised fashion to take advantage of the majority of the available data which is unlabeled and output a binary segmentation mask used to determine the localization and boundaries of a peak signal within an XIC, if there exists one.

To evaluate the performance of the proposed method and model architecture, we will benchmark them on the manually annotated subset of a representative LC-MS dataset against both baseline neural network architectures and current state-of-the-art peak detection systems. We find that our model outperforms the baseline architectures and is competitive with current state-of-the-art-systems.

## 2. METHOD

As shown in Fig. 2 (a), in order to take advantage of unlabeled data for training, we adapt FixMatch, a semisupervised image classification technique, for multivariate time series segmentation [15].

### 2.1. Augmentations

For augmentations, we utilize a collection of perturbations sourced from time series literature, speech recognition, and computer vision [7, 9, 20]. These include linear scaling of intensity values, jittering, shuffling of channels, and masking continuous stretches of the input along both axes. For our weak augmentations, only linear scaling is used, whereas all augmentations are applied with a 50% probability per sample for strong augmentations. We also implement a 1D mixing procedure for regularization based off CowMix [3], a variant of CutMix [20] designed for image segmentation tasks, where a binary mask is randomly generated and used to combine two unlabeled XICs.

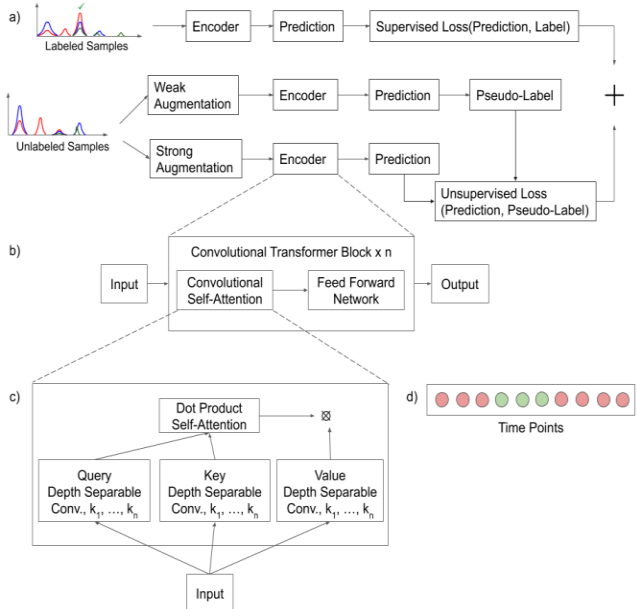

**Figure 2:** a) The FixMatch loss is the sum of a labeled loss and an unsupervised consistency loss generated by comparing the model output of weakly augmented inputs against pseudo-labels generated by the same model on strongly augmented inputs. The Conformer blocks in (b) use convolutional self-attention layers, c), where depth separable convolutions of *n* differently sized kernels transform the input into query, key, and value vectors that are aware of local context. The label vector of an XIC, d), consists of a binary vector with a peak (green)/non-peak (red) label for each individual time point.

### 2.2. Neural Network Architecture

A vanilla Transformer consists of positional embeddings and a stack of encoder blocks, which are themselves composed of a self-attention layer and a position-wise feed-forward network, regularized using layer normalization and residual dropout connections [18]. Since one channel of our input XICs contains relative position information, we omit the positional embedding layer. The self-attention layer utilizes a dot-product attention mechanism, which measures the similarity between a query and a key vector, and uses the similarity to output a weighted sum of a value vector that corresponds to the key:

$$\text{Attention}(Q, K, V) = \text{softmax}((QW^Q K^T W^K)/d_k)VK^V, \quad (1)$$

where Q, K, and V are the query, key, and value matrices, $W^{Q, K, V}$ are learnable projection matrices, and $d_k$ is the dimension of the key. In self-attention, Q == K == V.

For our proposed model, we chose to implement a Convolutional Transformer (i.e. a Conformer) as seen in Fig. 2 (c), where we replace the linear projections $W^{Q, K, V}$ in (1) with depth separable, gated kernel size convolutions [1]. For each of these, the input goes through a pointwise (1x1) convolution, and then a series of depth wise convolutions in parallel (by default using kernel sizes of 3 and 15 to cover both small- and large-scale features). The outputs are combined as a weighted sum parameterized by a learned gating vector, which allows each layer to select whether it wants to focus on fine grained or larger scale features. The goal of this convolutional self-attention is to overcome a CNN's equivariance to translation and allow the model to rank multiple detections in a single XIC, and to overcome the Transformer's lack of awareness of local context like shape during query-key matching [4, 18].

The position-wise feed-forward network is unchanged, and still applied at each position of the output of the self-attention layer. However, instead of layer normalization, since we do not want to normalize along the feature dimension, only time, we replace them with affine instance normalization [17]. Finally, we utilize an adaptive input normalization layer, so no further preprocessing of the input is necessary after XIC creation [8].

## 2.3. Loss

We use a weighted version of binary cross entropy called Focal Loss, which puts more weight on difficult examples [5]. Focal Loss also helps with class imbalances; there are frequently more labeled negative XICs than positive, and even on a pointwise level, peaks usually consist of a very small fraction of the total length of an XIC (e.g. 5 points out of 175, or less than 3% of the total length).

## 3. EXPERIMENTS

### 3.1. Dataset

We tested our method using a representative LC-MS dataset, containing 431,470 XICs from 16 equal sized LC-MS/MS runs, of which 7,103 were manually annotated with exact peak locations (originating from 452 distinct peptides), and 123,334 are synthetic negative XICs [13]. XICs were split into training (~70%), validation (~20%), and holdout test (~10%) sets.

### 3.2. Input Features

The raw two dimensional time series spectra can span across hundreds of m/z values and thousands of RT points (representing experiments of multiple hours, with each time point spaced ca. 3.4 seconds apart), reaching dimensions of upwards of 500,000 by 2,000 due to the high resolution of the data. To facilitate easier experimentation and analysis, we bin the data at the m/z axis at a resolution of 0.01 m/z per bin.

Instead of summing up values within m/z ranges for each fragment, we instead extract 70 bins of width 0.01 m/z for the precursor itself (from MS1) and each of its fragments (from MS2), across a window of 175 time points along the RT axis based on the expected RT of our library compounds (approx. 10 min. worth of data or 8% of the full experiment) from the raw data, similar to other deep learning works in the field of proteomics [16]. We treat this data as an n-channel time series, and add additional channels representing precursor specific information from the assay library, resulting in 498-channel XICs of length 175 (see Fig. 3) that are fed directly into our neural network. This allows us to maintain structural information along the m/z dimension otherwise lost in other peak detection methods. Note that for Fig. 1 the 70 m/z bins are aggregated into a single trace for ease of visualization, and for both Fig. 1 and Fig. 3 the additional information channels are omitted.

For our training labels, we convert the manual annotations into a target binary mask by inserting positive labels at each index which lies between the indices corresponding to the left and right boundaries of an annotation, as seen in Fig. 2 (d).

### 3.3. Experimental Configurations

The baseline CNN consisted of the adaptive normalization layer and 6 standard blocks consisting of a convolutional layer, 1D batch normalization layer, and a ReLU activation, starting with 256 convolutional filters and halving for each block, kernel sizes starting at 13 and decreasing by 2 for each block, same/zero padding, and a final linear classification layer with a sigmoid activation.

The baseline Transformer also included the adaptive normalization layer, 6 standard blocks consisting of a self-attention layer, skip connections, dropout with 10% probability, layer normalization, a position wise feedforward layer with an intermediate channel multiplier of 4 and ReLU activations, and the same final layer as the CNN. Each transformer block utilized a channel size of 64, and a single self-attention head. The Conformer shared the same parameter values as the Transformer.

All models were trained using the built in PyTorch implementations of the AdamW optimizer with a maximum learning rate of 0.003 and a weight decay of 0.3, and a cosine annealed learning rate scheduler [6, 14]. Any other parameter values not mentioned utilized the PyTorch defaults.

### 3.4. Output Post Processing and Evaluation Metrics

To obtain our final peak predictions, we take the pointwise output of the model and threshold it to obtain a binary segmentation mask. Then we extract all continuous positive regions of length 3 or more to represent the model's detections. The score of each detection is taken to be the maximum model output score within it. No further post processing is performed.

Models are evaluated using Average Precision (AP) at Intersection over Union (IoU) thresholds of 0.3:.05:0.7. Although a more lenient standard than commonly used in the 2D object detection community, for mass spectrometry data there are many cases where the human annotators over annotate peak signal boundaries to be wider than they really are. The range of IoU thresholds from 0.3 to 0.7 provides a good balance between being permissive enough for these cases as well as penalizing truly unacceptable localization errors.

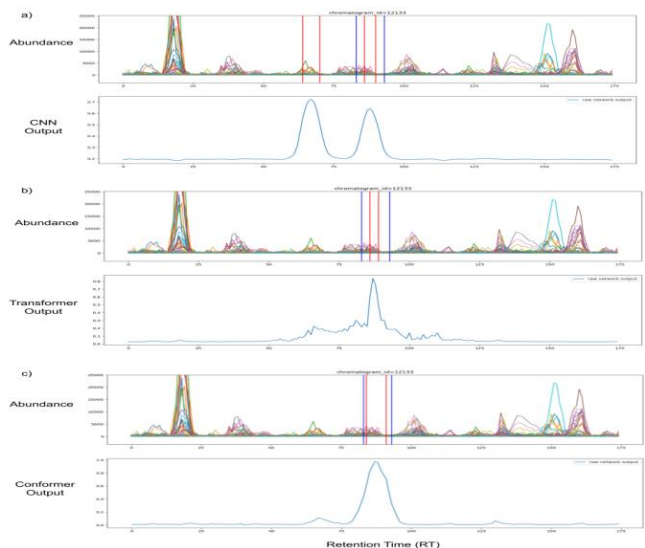

**Figure 3:** a), b), and c) show the raw network outputted pointwise probabilities of a CNN, Transformer, and Conformer network on the same XIC, respectively. We can observe common characteristics of each network type's output. Blue vertical lines denote the ground truth boundaries, and red vertical lines represent the model's predicted boundaries (points above 0.5).

### 3.5. Results and Discussion

We measured the performance of 5 models: a supervised baseline CNN, a supervised baseline Transformer, a supervised Conformer, a semisupervised Conformer, and a state-of-the-art system based on XGBoost, which is trained in a semisupervised manner [11]. The supervised models are trained on all manually annotated XICs in the full training set, as well as a proportional subset of the synthetic negative XICs randomly sampled from the total negative XICs in the training set such that the final ratio of positive to negative samples matches that of the original dataset. The validation and test sets for the supervised models are constructed in a similar manner. The semisupervised models are trained on all available samples in the training set, however their validation and test sets are kept the same as the supervised trials. The reported results in the table below were averaged over three experimental replicates for each model and evaluated on the subsetted holdout test data using the best performing models on the subsetted validation data. Models with a suffix of "-S" denote a supervised model, and a suffix of "-SS" denotes a semisupervised model.

First, we trained supervised versions of each architecture to determine their suitability for peak detection. The outputted binary segmentation masks are thresholded at the default value of 0.5. It can be seen from Table 1 that the Conformer architecture obtains the highest AP and AR scores in the supervised task, and the semisupervised training further improves its performance. We also note that the relative performance does not change with different IoU thresholds.

| Model | AP | AR |
|---|---|---|
| CNN-S | 87.36 | 58.67 |
| Transformer-S | 77.74 | 51.16 |
| Conformer-S | 90.20 | 66.13 |
| Conformer-SS | **90.73** | **72.12** |

**Table 1:** AP and AR scores obtained across IoU thresholds 0.3:0.05:0.7 by baseline neural architectures and the proposed model (in %). Highest metric values are bolded.

From Fig. 3 a), we can see that the CNN can have false detections when multiple possible peak signal regions exist. On the other hand, in Fig. 3 b), the Transformer has many more higher scoring points outside of annotated regions due relying on pointwise comparisons without leveraging local context.

Finally, we compare the performance of the final semisupervised Conformer with the current state-of-the-art, but only on their top detection per XIC (as in a real use case). For this comparison, we also tune the binary segmentation mask threshold value of the neural network output on the validation set and find that we can lower it to 0.35 from 0.5 while maintaining precision but greatly boosting recall. The threshold of the state-of-the-art system is automatically tuned on the validation set by its software package [11]. As shown in Table 2, we can achieve competitive performance to the current state-of-the-art with minimal feature engineering. We observed that the proposed model outperforms the state-of-the-art at higher IoU thresholds, and vice versa; this suggests that while the state-of-the-art may be the top choice for identification, the proposed model may be preferable for cases when high quality quantification is the goal.

| Model | AP | AR |
|---|---|---|
| Conformer-SS Tuned Threshold and Top 1 | **91.98** | 82.41 |
| XGBoost-SS Tuned Threshold and Top 1 | 89.38 | **83.99** |

**Table 2:** AP and AR scores obtained across IoU thresholds 0.3:0.05:0.7 on only the top detection per model based on a tuned threshold value by the proposed model and a state-of-the-art system (in %). Highest metric values are bolded.

### 4. CONCLUSION

This work presents a novel framework for peak detection in DIA-MS proteomics data; however our approach can operate on any set of multichannel time series-like input, such as other forms of chromatography based mass spectrometry, or even unrelated data such as audio spectrograms or multi-sensor outputs, and is easily extensible to include additional sources of information by simply appending additional channels to the input. Our Convolutional Transformer can overcome common challenges faced by other neural architectures for peak detection and outperforms other approaches on a manually annotated subset of a representative LC-MS dataset. We further boost performance by training on a much larger unlabeled dataset in a semi-supervised manner, which is important in real-world applications. Future work will attempt to further reduce or eliminate completely the reliance on manual labels, explore multi-stage training to separate the tasks of peak signal detection and boundary generation, and to incorporate additional dimensions of information, to further enhance both the quantitative and qualitative performance of our method.

## 5. REFERENCES

[1] F. Chollet, "Xception: Deep Learning with Depthwise Separable Convolutions," 2017 IEEE Conference on Computer Vision and Pattern Recognition (CVPR), 2017.

[2] V. B. Demichev, C. S. Messner, K. undefined Lilley, and M. undefined Ralser, "DIA-NN: Deep neural networks substantially improve the identification performance of Data-independent acquisition (DIA) in proteomics," Nature Methods, vol. 17, pp. 41–44, Nov. 2019.

[3] G. French, T. Aila, S. Laine, M. Mackiewicz, and G. Finlayson, "Semi-supervised semantic segmentation needs strong, high-dimensional perturbations," arXiv.org, 26-Sep-2019. [Online]. Available: https://arxiv.org/abs/1906.01916v2. [Accessed: 21-Oct-2020].

[4] S. Li, X. Jin, Y. Xuan, X. Zhou, W. Chen, Y.-X. Wang, and X. Yan, "Enhancing the Locality and Breaking the Memory Bottleneck of Transformer on Time Series Forecasting," arXiv.org, 03-Jan-2020. [Online]. Available: https://arxiv.org/abs/1907.00235. [Accessed: 21-Oct-2020].

[5] T.-Y. Lin, P. Goyal, R. Girshick, K. He, and P. Dollár, "Focal Loss for Dense Object Detection," arXiv.org, 07-Feb-2018. [Online]. Available: https://arxiv.org/abs/1708.02002. [Accessed: 21-Oct-2020].

[6] I. Loshchilov and F. Hutter, "Decoupled Weight Decay Regularization," arXiv.org, 04-Jan-2019. [Online]. Available: https://arxiv.org/abs/1711.05101. [Accessed: 21-Oct-2020].

[7] D. S. Park, W. Chan, Y. Zhang, C.-C. Chiu, B. Zoph, E. D. Cubuk, and Q. V. Le, "SpecAugment: A Simple Data Augmentation Method for Automatic Speech Recognition," Interspeech 2019, 2019.

[8] N. Passalis, A. Tefas, J. Kanniainen, M. Gabbouj, and A. Iosifidis, "Deep Adaptive Input Normalization for Time Series Forecasting," arXiv.org, 22-Sep-2019. [Online]. Available: https://arxiv.org/abs/1902.07892. [Accessed: 21-Oct-2020].

[9] K. M. Rashid and J. Louis, "Times-series data augmentation and deep learning for construction equipment activity recognition," Advanced Engineering Informatics, 27-Jun-2019. [Online]. Available: https://www.sciencedirect.com/science/article/pii/S1474034619300886. [Accessed: 21-Oct-2020].

[10] L. Reiter, O. Rinner, P. Picotti, R. Hüttenhain, M. Beck, M.-Y. Brusniak, M. O. Hengartner, and R. Aebersold, "mProphet: automated data processing and statistical validation for large-scale SRM experiments," Nature Methods, vol. 8, no. 5, pp. 430–435, 2011.

[11] G. Rosenberger, I. Bludau, U. Schmitt, M. Heusel, C. L. Hunter, Y. Liu, M. J. Maccoss, B. X. Maclean, A. I. Nesvizhskii, P. G. A. Pedrioli, L. Reiter, H. L. Röst, S. Tate, Y. S. Ting, B. C. Collins, and R. Aebersold, "Statistical control of peptide and protein error rates in large-scale targeted data-independent acquisition analyses," Nature Methods, vol. 14, no. 9, pp. 921–927, 2017.

[12] H. L. Röst, G. Rosenberger, P. Navarro, L. Gillet, S. M. Miladinović, O. T. Schubert, W. Wolski, B. C. Collins, J. Malmström, L. Malmström, and R. Aebersold, "OpenSWATH enables automated, targeted analysis of data-independent acquisition MS data," Nature Biotechnology, vol. 32, no. 3, pp. 219–223, 2014.

[13] H. L. Röst, Y. Liu, G. Dagostino, M. Zanella, P. Navarro, G. Rosenberger, B. C. Collins, L. Gillet, G. Testa, L. Malmström, and R. Aebersold, "TRIC: an automated alignment strategy for reproducible protein quantification in targeted proteomics," Nature Methods, vol. 13, no. 9, pp. 777–783, 2016.

[14] L. N. Smith and N. Topin, "Super-Convergence: Very Fast Training of Neural Networks Using Large Learning Rates," arXiv.org, 17-May-2018. [Online]. Available: https://arxiv.org/abs/1708.07120. [Accessed: 21-Oct-2020].

[15] K. Sohn, D. Berthelot, C.-L. Li, Z. Zhang, N. Carlini, E. D. Cubuk, A. Kurakin, H. Zhang, and C. Raffel, "FixMatch: Simplifying Semi-Supervised Learning with Consistency and Confidence," arXiv.org, 21-Jan-2020. [Online]. Available: https://arxiv.org/abs/2001.07685. [Accessed: 21-Oct-2020].

[16] N. H. Tran, R. Qiao, L. Xin, X. Chen, C. Liu, X. Zhang, B. Shan, A. Ghodsi, and M. Li, "Deep learning enables de novo peptide sequencing from data-independent-acquisition mass spectrometry," Nature Methods, vol. 16, no. 1, pp. 63–66, 2018.

[17] D. Ulyanov, A. Vedaldi, and V. Lempitsky, "Instance Normalization: The Missing Ingredient for Fast Stylization," arXiv.org, 06-Nov-2017. [Online]. Available: https://arxiv.org/abs/1607.08022. [Accessed: 21-Oct-2020].

[18] A. Vaswani, N. Shazeer, N. Parmar, J. Uszkoreit, L. Jones, A. N. Gomez, Ł. Kaiser, and I. Polosukhin, "Attention is all you need," Attention is all you need | Proceedings of the 31st International Conference on Neural Information Processing Systems, 01-Dec-2017. [Online]. Available: https://dl.acm.org/doi/10.5555/3295222.3295349. [Accessed: 21-Oct-2020].

[19] Z. Wu, D. Serie, G. Xu, and J. Zou, "PB-Net: Automatic peak integration by sequential deep learning for multiple reaction monitoring," Journal of Proteomics, 13-May-2020. [Online]. Available: https://www.sciencedirect.com/science/article/abs/pii/S1874391920301883?via=ihub. [Accessed: 21-Oct-2020].

[20] S. Yun, D. Han, S. Chun, S. J. Oh, Y. Yoo, and J. Choe, "CutMix: Regularization Strategy to Train Strong Classifiers With Localizable Features," 2019 IEEE/CVF International Conference on Computer Vision (ICCV), 2019.